\documentclass[letterpaper, 10 pt, conference]{ieeeconf}  % Comment this line out if you need a4paper

\IEEEoverridecommandlockouts                              % This command is only needed if 
\usepackage{siunitx}
\usepackage{hyperref}
\usepackage[T1]{fontenc}
\usepackage{booktabs}
\usepackage{graphicx}  % drop `demo` option in actual document!
\usepackage[]{subfig}
\usepackage{amsmath}
\usepackage{amssymb}
\usepackage{multirow}
\usepackage{amsfonts} % mathfrak
\usepackage{algorithm}
\usepackage[noend]{algpseudocode}
\makeatletter
\def\BState{\State\hskip-\ALG@thistlm}
\makeatother
\newcommand{\bigzero}{\mbox{\normalfont\Large\bfseries 0}}
\algnewcommand\algorithmicforeach{\textbf{for each}}
\algdef{S}[FOR]{ForEach}[1]{\algorithmicforeach\ #1\ \algorithmicdo}
\title{\LARGE \bf
Real-time marker-less multi-person 3D pose estimation in RGB-Depth camera networks %* #under Clutter and Occlusion
}

\author{Marco Carraro$^1$, Matteo Munaro$^1$, Jeff Burke$^2$ and Emanuele Menegatti$^1$% <-this % stops a space
\thanks{$^1$Marco Carraro, Matteo Munaro and Emanuele Menegatti are with the Intelligent Autonomous Systems Laboratory (IAS-Lab), Department of Information Engineering (DEI), University of Padova, Via Ognissanti 72, 35129, Padova, Italy. {\tt\small \{marco.carraro, emg\}@dei.unipd.it}}
\thanks{$^2$Jeff Burke is with REMAP in the School of Theater, Film and Television at UCLA, Los Angeles, California, USA 90095}
}

\begin{document}

\maketitle
%\begin{figure*}[t]
%\centering
%\dashbox{5}(400,100){\huge images}
%\caption{caption}
%\end{figure*}

\thispagestyle{empty}
\pagestyle{empty}

%%%%%%%%%%%%%%%%%%%%%%%%%%%%%%%%%%%%%%%%%%%%%%%%%%%%%%%%%%%%%%%%%%%%%%%%%%%%%%%%
\begin{abstract}

This paper proposes a novel system to estimate and track the 3D poses of multiple persons in calibrated RGB-Depth camera networks. 
The multi-view 3D pose of each person is computed by a central node which receives the single-view outcomes from each camera of the network. Each single-view outcome is computed by using a CNN for 2D pose estimation and extending the resulting skeletons to 3D by means of the sensor depth.
The proposed system is marker-less, multi-person, independent of background and does not make any assumption on people appearance and initial pose. The system provides real-time outcomes, thus being perfectly suited for applications requiring user interaction. 
Experimental results show the effectiveness of this work with respect to a baseline multi-view approach in different scenarios. To foster research and applications based on this work, we released the source code in OpenPTrack, an open source project for RGB-D people tracking.
\end{abstract}

%%%%%%%%%%%%%%%%%%%%%%%%%%%%%%%%%%%%%%%%%%%%%%%%%%%%%\%%%%%%%%%%%%%%%%%%%%%%%%%%%
\section{INTRODUCTION}
\label{sec:introduction}

The human body pose is rich of information. Many algorithms and applications, such as Action Recognition~\cite{han2017simultaneous, zanfir2013moving, wang2013approach}, People Re-identification~\cite{ghidoni2017multi}, Human-Computer-Interaction (HCI)~\cite{jaimes2007multimodal} and Industrial Robotics~\cite{morato2014toward,michieletto2017flexicoil,stival2017how} rely on this type of data. 
%to i) predict an action a person is going to do, ii) actively interact with humans or iii) build cooperative and safe human-robot assembly lines. 
The recent availability of smart cameras~\cite{zivkovic2010wireless, carraro2016powerful, carraro2016cost} and affordable RGB-Depth sensors as the first and second generation Microsoft Kinect, allow to estimate and track body poses in a cost-efficient way. However, using a single sensor is often not reliable enough because of occlusions and Field-of-View (FOV) limitations. For this reason, a common solution is to take advantage of camera networks. Nowadays, the most reliable way to perform human Body Pose Estimation (BPE) is to use marker-based motion capture systems. These systems show great results in terms of accuracy (less than 1mm), but they are very expensive and require the users to wear many markers, thus imposing heavy limitations to their diffusion. Moreover, these systems usually require offline computations in complicated scenarios with many markers and people, while the system  we propose provides immediate results. 
A real-time response is usually needed in security applications, where person actions should be detected in time, or in industrial applications, where human motion is predicted to prevent collisions with robots in shared workspaces.
Aimed by those reasons, the research on marker-less motion capture systems has been particularly active in recent years. 

%\begin{figure}[!t]
%\centering
%  %\includegraphics[width=\textwidth]{images/system_overview.png}
%  \includegraphics[width=1.0\linewidth]{images/Teaser_image.png}  
%  \caption{The output provided by the system we are proposing. In this example, two persons are seen from a network composed of 4 Microsoft Kinect v2.}
%\label{fig:teaser}
%\end{figure}
\begin{figure}[!t]
\centering
  \includegraphics[width=1.0\linewidth]{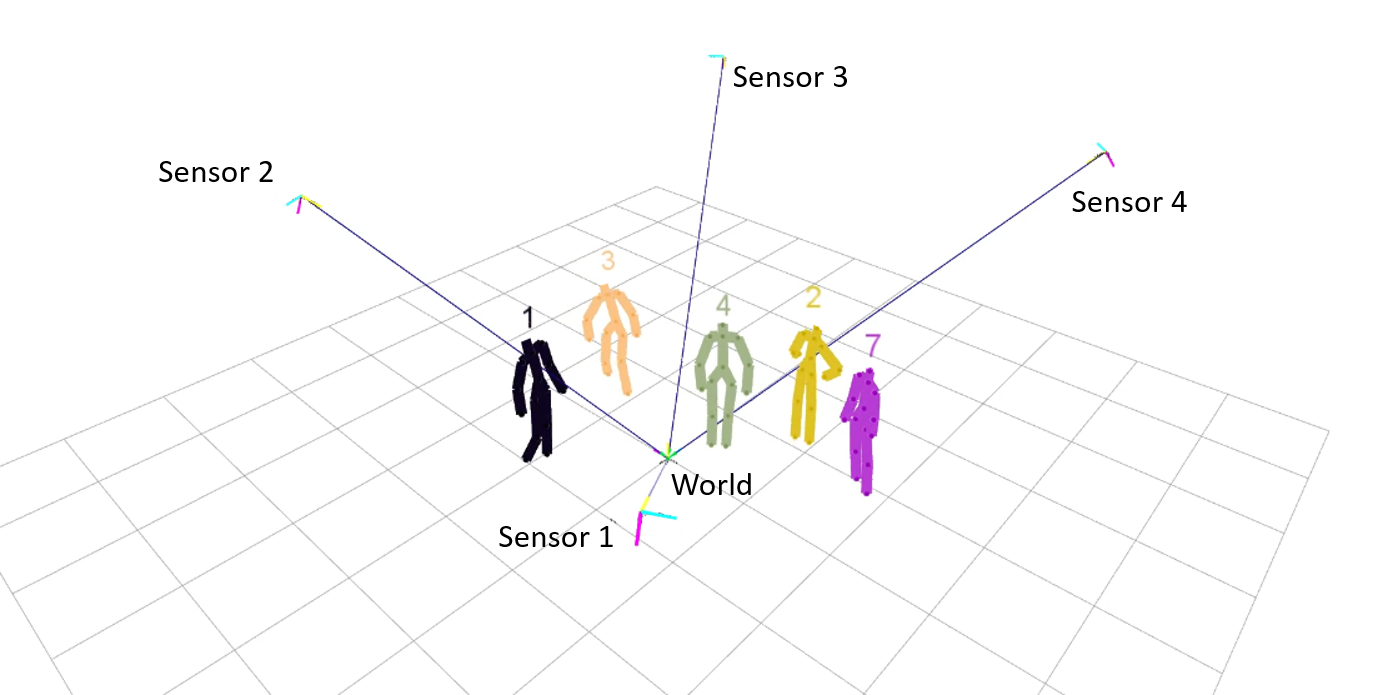}  
  \caption{The output provided by the system we are proposing. In this example, five persons are seen from a network composed of four Microsoft Kinect v2.}
\label{fig:teaser}
\end{figure}
\begin{figure*}[!t]
\centering
  \includegraphics[width=0.92\textwidth]{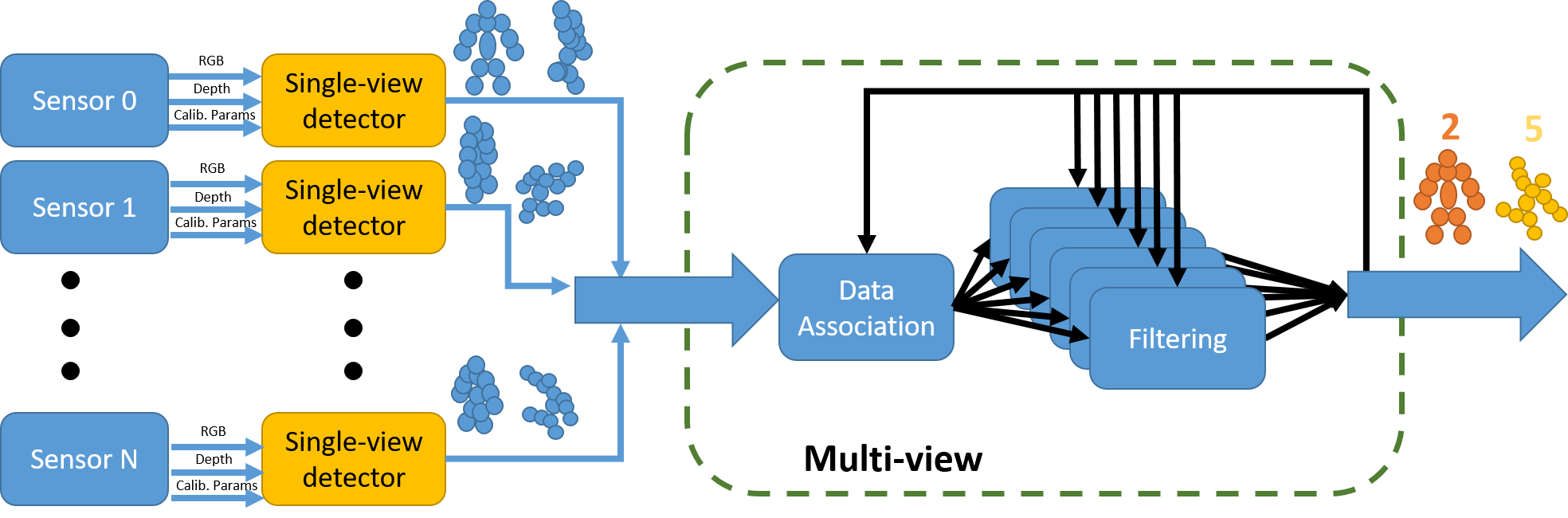}  
  \caption{The system overview. The camera network is composed of several RGB-D sensors (from 1 to N). Each single-view detector takes the RGB and Depth images as input and computes the 3D skeletons of the people in the scene as the output using the calibration parameters K. The information is then sent to the multi-view central node which is in charge of computing the final pose estimation for each person in the scene. First, a data association is performed to determine which pose detection is belonging to which pose track, then a filtering step is performed to update the pose track given the detection.}
  %The node also mantains a unique track ID for each person.
\label{fig:system}
\end{figure*}
In this work, we propose a novel system to estimate the 3D human body pose in real-time. To the best of our knowledge, this is the first open-source and real-time solution to the multi-view, multi-person 3D body pose estimation problem. Figure~\ref{fig:teaser} depicts our system output. The system relies on the feed of multiple RGB-D sensors (from 1 to N) placed in the scene and on an extrinsic calibration of the network. in this work, this calibration is performed with the \texttt{calibration\_toolkit}~\cite{basso2014online}\footnote{\url{https://github.com/iaslab-unipd/calibration\_toolkit}}. The multi-view poses are obtained by fusing the single view outcomes of each detector, that runs a state-of-the-art 2D body pose estimator~\cite{wei2016cpm, cao2017realtime} and extend it to 3D by means of the sensor depth.
The contribution of the paper is two-fold: i) we propose a novel system to fuse and update 3D body poses of multiple persons in the scene and ii) we enriched a state-of-the-art single-view 2D pose estimation algorithm to provide 3D poses. As a further contribution, the code of the project has been released as open-source as part of the OpenPTrack~\cite{munaro2014openptrack, munaro2016openptrack} repository.
%, together with the documentation and some ready-to-run examples. 
The proposed system is:
\begin{itemize}
\item \textit{multi-view}: The fused poses are computed taking into account the different poses of the single-view detectors;
\item \textit{asynchronous}: The fusion algorithm does not require the different sensors to be synchronous or have the same frame rate. This allows the user to choose the detector computing node accordingly to his needs and possibilities;
\item \textit{multi-person}: The system does not make any assumption on the number of persons in the scene. The overhead due to the different number of persons is negligible;
\item \textit{scalable}: No assumptions are made on the number or positions of the cameras. The only request is an offline one-time extrinsic calibration of the network;
%\item \textit{model-free}: We are not using any model to compute our poses;
\item \textit{real-time}: The final pose framerate is linear to the number of cameras in the network. In our experiments, a single-camera network can provide from 5 fps to 15 fps depending on the Graphical Processing Unit (GPU) exploited by the detector. The final framerate of a camera network composed of $k$ nodes is the sum of their single-view framerate; 
\item \textit{low-cost}: The system relies on affordable low-cost RGB-D sensors controlled by consumer GPU-enabled computers. No specific hardware is required. 
%For this reason, we believe it can really be used as the \textit{poor-man} solution to common motion-capture systems;
\end{itemize}

The remainder of the paper is organized as follows: in
Section~\ref{sec:related_work} we review the literature regarding human BPE from single and multiple views, while Section~\ref{sec:system_design} describes our system and the approach used to solve the problem. In Section~\ref{sec:experiments} experimental results are presented, and, finally in Section~\ref{sec:conclusions} we present our final conclusions.

\section{RELATED WORK}
\label{sec:related_work}

%Human body pose estimation is a broad field, here we review the works that are similar to ours, i.e. they use camera networks and they do not require any type of marker. Since many state-of-the-art works combine single-view pose estimators we will also review this category.

\subsection{Single-view body pose estimation}
Since a long time, there have been a great interest about single-view human BPE, in particular for gaming purposes or avatar animation. Recently, the advent of affordable RGB-D sensors boosted the research in this and other Computer Vision fields.
Shotton et al.~\cite{shotton2013real} proposed the skeletal tracking system licensed by Microsoft used by the XBOX console with the first-generation Kinect. This approach used a random forest classifier to classify the different pixels as belonging to the different body parts. This work inspired an open-source approach that was released by Buys et al.~\cite{buys2014adaptable}. This same work was then improved by adding the OpenPTrack people detector module as a preprocessing step~\cite{carraro2016improved}. Still, the performance of the detector remained very poor for non frontal persons. 
In these last years, many challenging Computer Vision problems have been finally resolved by using \textit{Convolutional Neural Networks} (CNNs) solutions. Also single-view BPE has seen a great benefit from these techniques~\cite{insafutdinov2016deepercut, pishchulin2016deepcut, carreira2016human, cao2017realtime}. The impressive pose estimation quality provided by those solution is usually paid in terms of computational time. Nevertheless, this limitation is going to be leveraged with newer network layouts and Graphical Processing Units (GPU) architectures, as proved by some recent works~ \cite{carreira2016human, cao2017realtime}. In particular, the work of Cao et. al~\cite{cao2017realtime} was one of the first to implement a CNN solution to solve people BPE in real-time using a bottom-up approach. The authors were able to compute 2D poses for all the people in the scene with a single forward pass of their CNN. This work has been adopted here as part of our single-view detectors.
%\cite{yang2013articulated}\cite{yang2011articulated}, 

\begin{figure*}[tbp]
\centering
  \includegraphics[width=0.89\textwidth]{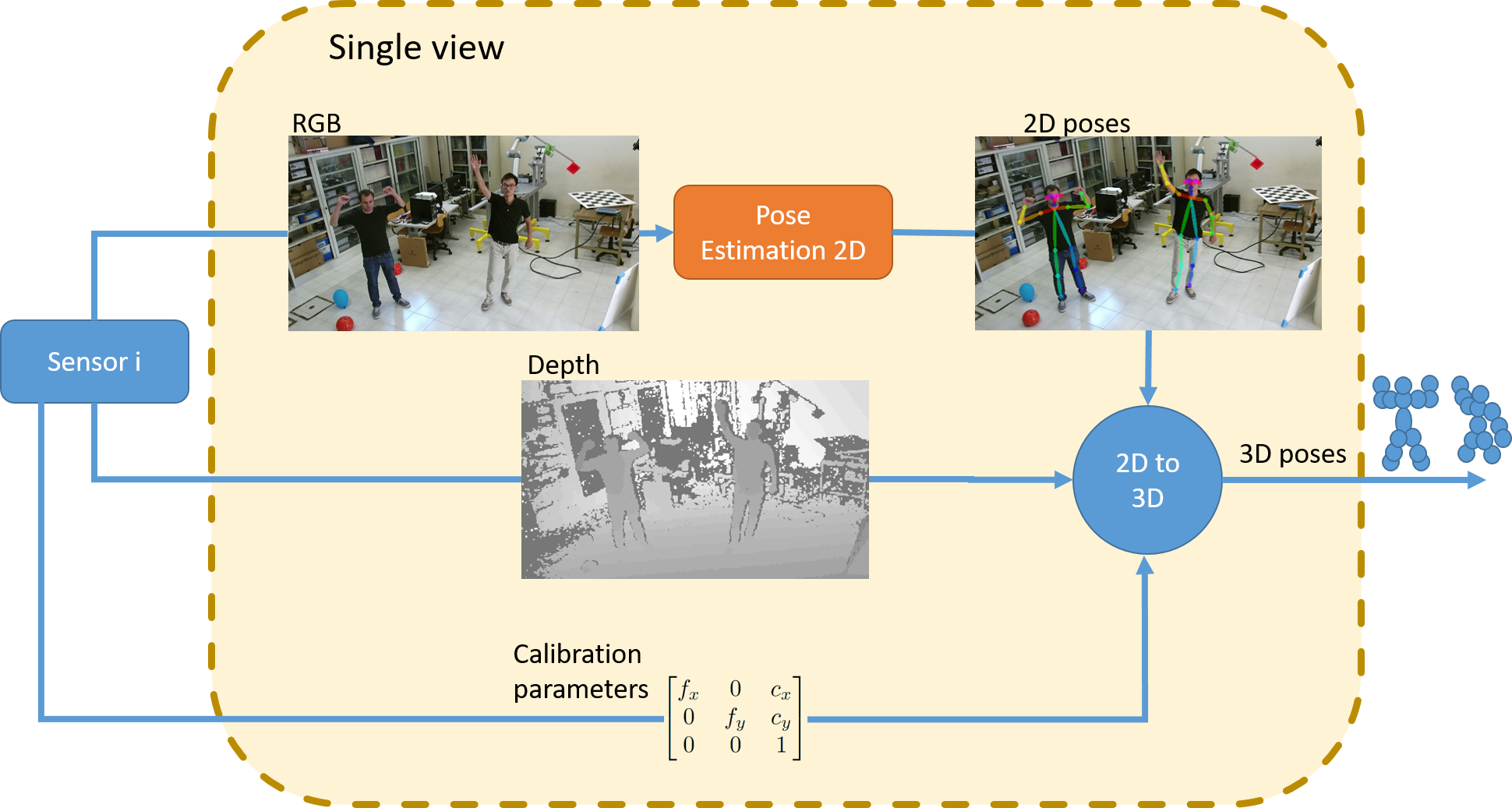}  
  \caption{The single-view pipeline followed for each sensor. At each new frame composed of a color image (RGB), a depth image and the calibration parameters, the 3D pose of each person in the scene is computed from the 2D one. Then, the results are sent to the central computer which will compute the multi-view result.}
\label{fig:single_view}
\end{figure*}

\subsection{Multi-view body pose estimation}
Multiple views can be exploited to be more robust against occlusions, self-occlusions and FOV limitations. 
%When multiple views are available, multiple problems  are  in generative and discriminative approaches. The first category is about works that compute the final pose of each persons by fitting a a human model to the input data from the different views. The second category usually implements Machine Learning solutions to directly infer the poses. 
In~\cite{elhayek2017marconi} a Convolutional Neural Network (CNN) approach is proposed to estimate the body poses of people by using a low number of cameras also in outdoor scenarios. The solution combines a generative and discriminative approach, since they use a CNN to compute the poses which are driven by an underlying model. For this reason, the collaboration of the users is required for the initialization phase.  
In our previous work~\cite{carraro2016improved}, we solved the single-person human BPE by fusing the data of the different sensors and by applying an improved version of~\cite{buys2014adaptable} to a virtual depth image of the frontalized person. In this way, the skeletonization is only performed once, on the virtual depth map of the person in frontal pose.
In~\cite{gao2015leveraging}, a 3D model is registered to the point clouds of two Kinects. The work provides very accurate results, but it is computationally expensive and not scalable to multiple persons.
The authors of \cite{lora2015geometric} proposed a pure geometric approach to infer the multi-view pose from a synchronous set of 2D single-view skeletons obtained using \cite{yang2013articulated}. The third dimension is computed by imposing a set of algebraic constraints from the triangulation of the multiple views. The final skeleton is then computed by solving a least square error method. While the method is computationally promising (skeleton computed in 1s per set of synchronized images with an unoptimized version of the code), it does not scale with the number of persons in the scene. 
In \cite{kim2017dance} a system composed of common RGB cameras and RGB-D sensors are used together to record a dance motion performed by a user. The fusion method is obtained by selecting the best skeleton match between the different ones obtained by using a probabilistic approach with a particle filter. The system performs well enough for its goal, but it does not scale to multiple people and requires an expensive setup. 
In \cite{kanaujia20113d} the skeletons obtained from the single images are enriched with a 3D model computed with the visual hull technique.
In \cite{yeung2013improved} two orthogonal Kinects are used to improve the single-view outcome of both sensors. They used a constrained optimization framework with the bone lengths as hard constraints. While the work provides a real-time solution and there are no hard assumption on the Kinect positions, it was tested just with one person and two orthogonal Kinect sensors.
Similarly to many recent works~\cite{lora2015geometric, kanaujia20113d, kim2017dance}, we use a single-view state-of-the-art body pose estimator, but we augment this result with 3D data and we then combine the multiple views to improve the overall quality. 
%To the best of our knowledge, this is the first work to provide a real-time, multi-person body pose estimation system that does not require any synchronization between the different imagers completely scalable in the number of people and cameras.

\section{SYSTEM DESIGN}
\label{sec:system_design}

Figure \ref{fig:system} shows an overview of the proposed system. 
%This is composed of several RGB-D sensors (detectors) and a central node called master computer where the multi-view part of the algorithm is computed. 
%For the sake of simplicity, we here consider a network composed of two sensors, but the extension to multiple sensors is trivial. 
It can be split into two parts: i) the single view, which is the same for each sensor and it is executed locally and ii) the multi-view part which is executed just by the master computer. In the single-view part (see Figure~\ref{fig:single_view}), each detector estimates the 2D body pose of each person in the scene using an open-source state-of-the-art single-view body pose estimator. In this work, we use the OpenPose\footnote{\url{https://github.com/CMU-Perceptual-Computing-Lab/openpose}}\cite{wei2016cpm,cao2017realtime} library, but the overall system is totally independent of the single-view algorithm used. The last operation made by the detector is to compute the 3D positions of each joint returned by OpenPose. This fusion is done by exploiting the depth information coming from the RGB-D sensor used. The 3D skeleton is then sent to the master computer for the fusion phase. This is done by means of multiple Unscented Kalman Filters used on the detection feeds, as explained in Section~\ref{subsec:multi-view}.

\subsection{Camera Network setup}
\label{subsec:camera-network-setup}

The camera network can be composed of several RGB-D sensors. In order to know the relative position of each camera, we calibrate the system using a solution similar to our previous works \cite{munaro2016openptrack, munaro2014openptrack}. From this passage we fix a common \textit{world} reference frame $\mathcal{W}$ and we obtain a transformation $\mathcal{T_{C}^\mathcal{W}}$, for each camera $C$ in the network, which transforms points in the camera coordinate system to the \textit{world} reference system.

\subsection{Single-view Estimation of 3D Poses}
\label{subsec:pose-estimation}

\begin{figure}[tbp]
\centering
  \includegraphics[width=0.8\linewidth]{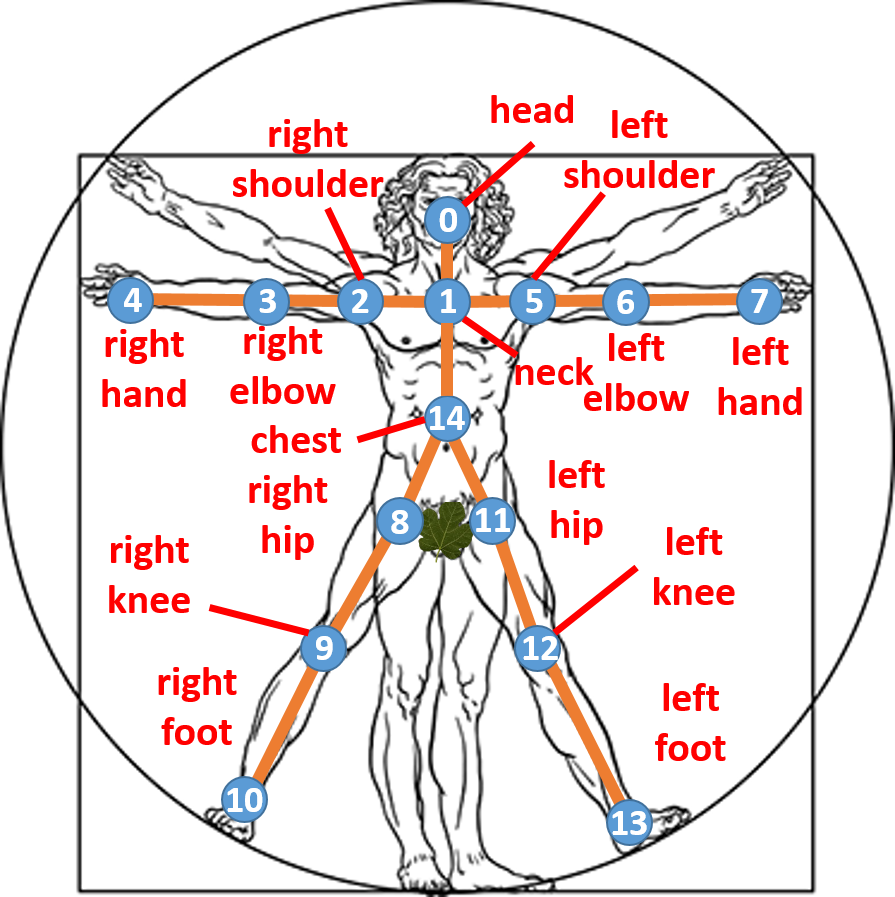}  
  \caption{The human model used in this work.}
\label{fig:human_model}
\end{figure}

\begin{algorithm}
\caption{The algorithm performed by the master computer to decide the association between the different skeletons in a detection and the current tracks.}\label{alg:data_association}
\begin{flushleft}
        \textbf{INPUT:} 
        \begin{itemize}
        \item $\mathfrak{^\mathcal{W}\widehat{S}_i} = \{S_0, S_1, ..., S_{k-1}\}$ - a new detection set from sensor $i$ in the world reference frame
        \item $\mathfrak{T} = \{T_0, T_1, ..., T_{l - 1}\}$ - the current set of tracked persons pose. 
        \item {$\epsilon $ - maximum distance for a detection to be considered for the association}
        \end{itemize}
        \textbf{OUTPUT:} 
        \begin{itemize}
        \item $\mathcal{M} = \{(S_i, T_j) \in       ^\mathcal{W}\mathfrak{\widehat{S}_i} \times \mathfrak{T} \}$ - the association between the pose tracked and the new observations
        \item $\mathcal{N} \subseteq ^\mathcal{W}\mathfrak{\widehat{S}_i}$ - the detections without an association. They will initialize a new track.
        \item $\mathfrak{T}_o \subseteq \mathfrak{T}$ - the tracks without an associated observations. They will be considered for removal
        \end{itemize}
\end{flushleft}
\begin{algorithmic}[1]
\Procedure{DATA\_ASSOCIATION}{$^\mathcal{W}\mathfrak{\widehat{S}_i}$, $\mathfrak{T}, \epsilon$}
\State {$\mathfrak{T}_o \gets \emptyset$}
\State {$C \gets \bigzero_{k \times l}$}
\ForEach {$T_i \in \mathfrak{T}$}
	\ForEach {$S_j \in ^\mathcal{W}\mathfrak{\widehat{S}_i}$}
		\label{al:line:centroid}
		\State{$x_t(j) \gets centroid(S_j)$}
		\label{al:line:z_t}
		\State {$z_t(i, j) \gets $ *v that $T_i$ would have if $S_j$ were associated to it*}
		\label{al:line:widehatz_t}
		\State {$\widehat{z}_{t|t-1}(i) \gets$ *prediction step of $\mathcal{K}_{im}$*}
		\label{al:line:S}
		\State {$\Sigma_t(i) \gets \Sigma_t(\mathcal{K}_{im})$}
		\label{al:line:tildez_t}
		\State {$\tilde{z}_t(i, j) = z_k(i, j) - \widehat{z}_{t|t-1}(i)$}
		\label{al:line:C}
		\State {$C_{ij} \gets \tilde{z}_t^T(i, j) \cdot \Sigma_t(i)^{-1} \cdot \tilde{z}_t(i, j)$}
	\EndFor
\EndFor
\label{al:line:munkres}
\State{$X \gets solve\_Munkres(C)$}
\For {$i \in [0, l - 1]$}
	\For {$j \in [i + 1, k - 1]$}
		\If {$X_{ij} == 1$ and $C_{ij} < \epsilon$}
			\label{al:line:M}
     		\State {$\mathcal{M} \gets \mathcal{M} \cup \{(S_j, T_i)\}$}
     		\State {* update $\mathcal{K}_{im}$ with $S_j$ *}
		\EndIf
	\EndFor
\EndFor
\label{al:line:N}
\State{$\mathcal{N} \gets \{S_i \, | \, \nexists T_j, \, (S_i, T_j) \in \mathcal{M}\}$}
\label{al:line:T}
\State{$\mathfrak{T}_o \gets \{T_i \, | \, \nexists S_j, \, (S_j, T_i) \in \mathcal{M}\}$}
\State {\Return $\mathcal{M}$, $\mathcal{N}$, $\mathfrak{T_o}$ }
\EndProcedure
\end{algorithmic}
\end{algorithm}
Each node in the network is composed of an RGB-D sensor and a computer to elaborate the images. Let $\mathfrak{^{R}F} = \{^{R}C, ^{R}D\}$ be a frame captured by the detector $R$ and composed of the color image $\textit{C}$ and the depth image $\textit{D}$ all in the $R$ reference frame. The color and depth images in $\mathfrak{F}$ are considered as synchronized. We then apply \textit{OpenPose} to $^RC$ obtaining the raw two dimensional skeletons $\overline{\mathfrak{S}} = \{\overline{S_0}, \overline{S_1}, ..., \overline{S_k}\}$. 
Each $S = \{ j_i \, | \, 0 \leq i \leq m\} \in \mathfrak{\overline{S}}$ is a set of 2D joints which follows the human model depicted in Figure~\ref{fig:human_model}.
The goal of the single-view detector is to transform $\overline{\mathfrak{S}}$ in the set of skeletons $\widehat{\mathfrak{S}} = \{\widehat{S_0}, \widehat{S_1}, ..., \widehat{S_k}\}$ where each $\widehat{S} \in \mathfrak{S}$ is a three dimensional skeleton. Given the RGB image $I$, let's consider a point $p = (x_p, y_p) \in I$  and its corresponding depth $d = proj(x_p, y_p)$. Considering $(f_x,f_y)$ and $(c_x, c_y)$ respectively the focal point and the optical center of the sensor, the relationship to compute the 3D point $P_R = (X_R, Y_R, Z_R)$ in the camera reference system R is explained in Equation~\ref{eq:pinhole_model}.
\begin{equation}
\label{eq:pinhole_model}
p = \begin{bmatrix}
        x_p \\
        y_p \\
        d 
	  \end{bmatrix} 
	= \begin{bmatrix}
	    f_x & 0   & c_x \\
	    0   & f_y & c_y \\
	    0   & 0   & 1   
	  \end{bmatrix}
	  \begin{bmatrix}
	    X_R \\
	    Y_R \\
	    Z_R
	  \end{bmatrix}
	= KP_R
\end{equation}  
Since the depth data is potentially noisy or missing, we compute the depth $d$ associated to the point $p = (x_p, y_p)$ by applying a median to the set $\mathfrak{D}(p)$, as shown in Equations~\ref{eq:neighborhood},~\ref{eq:2d_to_3d}. 
\begin{equation}
\centering
\label{eq:neighborhood}
	\mathfrak{D}(p=(x_p, y_p)) = \{ (x,y) \, | \, || (x,y) - (x_p, y_p)|| < \epsilon \}
\end{equation}
\begin{equation}
	\centering
	\label{eq:2d_to_3d}
	d = \phi(p) = \text{median}\{ proj(x,y) \, | \, (x,y) \in \mathfrak{D}(p)\}
\end{equation}
Given $\overline{\mathfrak{S}}$, we then proceed to the calculation of $\widehat{\mathfrak{S}}$ as shown in Equation~\ref{eq:2d_skel_to_3d_skel}.
\begin{align}
	\centering
	\label{eq:2d_skel_to_3d_skel}
\begin{split}
	\forall 0 & \leq j < k, \quad \overline{S_j} = \{\overline{j_i} = (x_i, y_i) \, | \, 0 \leq i < m\} \in \overline{\mathfrak{S}}, \\
	\widehat{S_j} & = \left\{\widehat{j_i} = 
	\begin{bmatrix}
		|K^{-1}(\overline{j_i})|_x \\
		|K^{-1}(\overline{j_i})|_y \\
		\phi(\overline{j_i})
	\end{bmatrix} \, , \, 0 \leq i < m\right\} \in \widehat{\mathfrak{S}} 
\end{split}
\end{align}

\begin{table*}[]
\centering
\resizebox{\textwidth}{!}{%
\begin{tabular}{clllllllllllll}
\hline
\multicolumn{1}{l}{}                                        &                                    & r-shoulder                & r-elbow                   & r-wrist                   & l-shoulder                & l-elbow                   & l-wrist                   & r-hip                     & r-knee                    & r-ankle                   & l-hip                     & l-knee                    & l-ankle                   \\ \hline
\multicolumn{1}{c|}{\multirow{3}{*}{single-camera network}} & \multicolumn{1}{l|}{$MAF_{30}$}   & \textgreater100          & \textgreater100          & \textgreater100          & \textgreater100          & \textgreater100          & \textgreater100          & \textgreater100          & \textgreater100          & \textgreater100          & \textgreater100          & \textgreater100          & \textgreater100          \\
\multicolumn{1}{c|}{}                                       & \multicolumn{1}{l|}{$MAF_{40}$}   & \textgreater100          & \textgreater100          & \textgreater100          & \textgreater100          & \textgreater100          & \textgreater100          & \textgreater100          & \textgreater100          & \textgreater100          & \textgreater100          & \textgreater100          & \textgreater100          \\
\multicolumn{1}{c|}{}                                       & \multicolumn{1}{l|}{\textbf{Ours}} & \textbf{54.9 $\pm$ 58.6} & \textbf{42.4 $\pm$ 47.4} & \textbf{42.4 $\pm$ 40.0} & \textbf{77.7 $\pm$ 74.4} & \textbf{79.1 $\pm$ 82.7} & \textbf{70.0 $\pm$ 61.8} & \textbf{51.7 $\pm$ 43.7} & \textbf{54.5 $\pm$ 31.0} & \textbf{63.3 $\pm$ 34.2} & \textbf{97.8 $\pm$ 30.3} & \textbf{57.5 $\pm$ 38.9} & \textbf{69.2 $\pm$ 37.6} \\ \hline
\multicolumn{1}{c|}{\multirow{3}{*}{2-camera network}}      & \multicolumn{1}{l|}{$MAF_{30}$}   & 62.0 $\pm$ 33.0          & 62.9 $\pm$ 32.0          & 63.1 $\pm$ 34.5          & 83.3 $\pm$ 33.4          & 85.8 $\pm$ 37.8          & 94.8 $\pm$ 45.4          & 76.4 $\pm$ 30.6          & 75.9 $\pm$ 27.4          & 88.3 $\pm$ 35.6          & \textgreater100         & 85.4 $\pm$ 35.5          & 93.3 $\pm$ 37.0          \\
\multicolumn{1}{c|}{}                                       & \multicolumn{1}{l|}{$MAF_{40}$}   & 83.7 $\pm$ 41.8          & 84.0 $\pm$ 40.9          & 83.1 $\pm$ 43.7          & \textgreater100         & \textgreater100 & \textgreater100        & 99.2 $\pm$ 40.4          & 96.3 $\pm$ 38.0          & \textgreater100         & \textgreater100 & \textgreater100         & \textgreater100         \\
\multicolumn{1}{c|}{}                                       & \multicolumn{1}{l|}{\textbf{Ours}} & \textbf{20.7 $\pm$ 17.2} & \textbf{21.0 $\pm$ 17.5} & \textbf{24.3 $\pm$ 17.5} & \textbf{32.1 $\pm$ 23.0} & \textbf{33.4 $\pm$ 26.3} & \textbf{39.8 $\pm$ 35.1} & \textbf{22.4 $\pm$ 16.7} & \textbf{42.8 $\pm$ 17.2} & \textbf{59.7 $\pm$ 28.6} & \textbf{98.3 $\pm$ 21.2} & \textbf{39.9 $\pm$ 18.3} & \textbf{58.6 $\pm$ 27.1} \\ \hline
\multicolumn{1}{c|}{\multirow{3}{*}{4-camera network}}      & \multicolumn{1}{l|}{$MAF_{30}$}   & 28.7 $\pm$ 16.4          & 31.0 $\pm$ 16.9          & 32.2 $\pm$ 22.5          & 41.5 $\pm$ 17.9          & 39.9 $\pm$ 19.6          & 44.7 $\pm$ 29.5          & 40.2 $\pm$ 15.0          & 48.7 $\pm$ 12.8          & 58.6 $\pm$ 21.2          & 94.1 $\pm$ 26.1          & 52.1 $\pm$ 17.8          & 57.8 $\pm$ 27.9          \\
\multicolumn{1}{c|}{}                                       & \multicolumn{1}{l|}{$MAF_{40}$}   & 38.4 $\pm$ 21.2          & 40.8 $\pm$ 21.7          & 41.6 $\pm$ 26.3          & 53.0 $\pm$ 23.2          & 52.7 $\pm$ 24.6          & 57.6 $\pm$ 33.1          & 50.7 $\pm$ 19.4          & 56.2 $\pm$ 16.7          & 66.0 $\pm$ 24.5          & 96.6 $\pm$ 30.8          & 61.2 $\pm$ 23.1          & 67.5 $\pm$ 31.6          \\
\multicolumn{1}{c|}{}                                       & \multicolumn{1}{l|}{\textbf{Ours}} & \textbf{22.7 $\pm$ 18.9} & \textbf{21.3 $\pm$ 18.5} & \textbf{26.3 $\pm$ 19.9} & \textbf{22.5 $\pm$ 22.1} & \textbf{26.7 $\pm$ 25.9} & \textbf{31.8 $\pm$ 29.7} & \textbf{23.9 $\pm$ 18.0} & \textbf{46.5 $\pm$ 19.7} & \textbf{55.9 $\pm$ 25.1} & \textbf{95.4 $\pm$ 22.0} & \textbf{45.1 $\pm$ 20.5} & \textbf{49.1 $\pm$ 25.2} \\ \hline
\end{tabular}
}
\caption{The results of the experiments. Each number represents the mean and the standard deviation of the reprojection error on a reference camera (see Equation~\ref{eq:reprerr})}.
\label{tab:results}
\end{table*}

\subsection{Multi-view fusion of 3D poses}
\label{subsec:multi-view}
The master computer is in charge of fusing the different information it is receiving from the single-view detectors in the network. One of the common limitations in motion capture systems is the necessity to have synchronized cameras. Moreover, off-the-shelves RGB-D sensors, such as the Microsoft Kinect v2, do not have the possibility to trigger the image acquisition. In order to overcome this limitation, our solution merges the different data streams asynchronously. This allows the system to work also with other RGB-D sensors or other low-cost embedded machine. 
At time $t$, the master computer maintains a set of tracks $\mathfrak{T} = \{T_0, T_1, ..., T_l\}$ where each pose tracked $T_i$ is composed of the set of states of $m$ different Kalman Filters, one per each joint, i.e: $T_i = \{\mathcal{S}(\mathcal{K}_{i0}), \mathcal{S}(\mathcal{K}_{i1}), ..., \mathcal{S}(\mathcal{K}_{im}) \}$. The additional Kalman Filter $\mathcal{K}_{im}$ is mantained for the data association algorithm. At time $t+1$, it may arrive a detection $\mathfrak{\widehat{S_i}} = \{\widehat{S_0}, \widehat{S_1}, ..., \widehat{S_k}\}$ from the sensor $i$ of the network. The master computer first refers the detection to the common \textit{world} coordinate system $\mathcal{W}$ (see Section \ref{subsec:camera-network-setup}):
\begin{equation*}
^\mathcal{W}\mathfrak{\widehat{S_i}} = \mathcal{T}_i^\mathcal{W} \cdot \mathfrak{\widehat{S_i}} = \{\mathcal{T}_i^\mathcal{W}\cdot S_j \, | \, \forall S_j \in \mathfrak{\widehat{S_i}}\}
\end{equation*}
Then, it associates the different skeletons in $^W\mathfrak{\widehat{S_i}}$ as new observations for the different tracks in $\mathfrak{T}$ if they belong to them or initializes new tracks if some of the skeletons do not belong to any $T_i \in \mathfrak{T}$. At this stage, the system also decides if a track is old and has to be removed from $\mathfrak{T}$. This step is important to prevent $\mathfrak{T}$ to grow big causing time computing problems with systems which are running for hours. We refer to this phase as \textit{data association}. Algorithm~\ref{alg:data_association} shows how it is performed.
The data association is done by considering the centroid of each skeleton $S$ contained in the detection $^\mathcal{W}\mathfrak{S_i}$. The centroid is calculated as the chest joint $j_{14} \in S$, if this is valid, otherwise it is replaced with a weighted mean of the neighbor joints. Lines~[\ref{al:line:z_t}-\ref{al:line:tildez_t}] of Algorithm~\ref{alg:data_association} refers to the calculation of a cost associated to the case if the detection pose $S_j$ would be associated to the track $T_i$. To calculate this, we consider the Mahalanobis distance between the likelihood vector at time $t$ $\tilde{z}_t(i,j)$ and $\Sigma_t(\mathcal{K}_{i,x_t})$: the covariance matrix of the Kalman filter associated to the centroid of $T_i$. At this point, computing the optimal association between tracks and detections is the same as solving the Hungarian algorithm associated to the cost matrix $C$; Line~\ref{al:line:munkres} refers to the use of the Munkres algorithm which efficiently computes the optimal matrix $X$ with a $1$ on the associated couples. Nevertheless, this algorithm does not consider a maximum distance between tracks and detections. Thus, it may happen that a couple is wrongly associated in the optimal assignment. For this reason, when inserting the couples in $\mathcal{M}$, we check also if the cost of the couple in the initial cost matrix $C$ is below a threshold.

Once solved the data association problem, we can assign the tracks ID to the different skeletons. Indeed, we know which are the detection at the current time $t$ belonging to the tracks in the system and, additionally, we know also which tracks need to be created (i.e. new detections with no associated track) and the tracks to consider for the removal. Let $n$ be the number of people in the scene, we used a set of Unscented Kalman Filters $\mathfrak{K} = \{\mathcal{K}_{ij}, \, 0 \leq i < n, \, 0 \leq j \leq m\}$ where the generic $\mathcal{K}_{ij} \in \mathfrak{K}$ is in charge of computing the new position of the joint $j$ of the person $i$ at time $t$, given the new detection received from one of the detectors at time $t$ and the prediction of the filter $\mathcal{K}_{ij}$ computed from the previous position at time $t - 1$ of the same joint $j$. 
%The first operation done by the master computer is the data association. The approach done in this work is novel with respect to our previous work. The data association is important, since at the generic time $t$ each $K_{ij} \in \mathfrak{K}$ needs the new joint $j$ of the person $k$. The data stream coming from the detectors is potentially un-ordered, since they do not have the notion of tracks, but just of people in the current frame. Therefore, in this frame, the master node must be able to associate the detection data to the correct track. Let $\mathfrak{S} = \{S_i, \, 0 \leq i < n\}$ be the set of skeleton tracks in the scene at time $t$, where each $S_i = \{S(K_i0), S(K_i1), ..., S(K_in)\} \in \mathfrak{S}$ is obtained by the set of the Kalman Filters states at time $t$ belonging to the person $i$, we used the 
%% TODO: Mahalanobis distance (se Munaro's article to put some information here

The state of each Kalman Filter $\mathcal{K}_{ij}$ is dimensioned with the three dimensional position of the joint $j$. We used as motion model a constant velocity model, since it is good to predict joint movements in the small temporal space between two good detections of that joint. 

\section{EXPERIMENTS}
\label{sec:experiments}

The algorithm described in this paper does not require any synchronization between the cameras in the networks.
This fact makes particularly difficult to find a fair comparison between our proposed system and other state-of-the-art works. %Moreover, it is particularly difficult to find public RGB-Depth multi-view datasets useful to perform such comparison. 
Thus, in order to provide useful indication on how our system performs, we recorded and manually annotated a set of RGB-D frames while a person was freely moving in the field-of-view of a 4-sensors camera network. 
We compare our algorithm with a baseline method called MAF (Moving Average Filter), in which the outcome of the generic joint $i$ at time $t$ is computed as an average of the last $k$ frames. In order to be as fair as possible, we fixed $k \geq 30$ to provide comparable results in terms of smoothness.
We also demonstrated the effectiveness of the multi-view fusion by comparing our results with the poses obtained by considering just one and two cameras of the same network. 
In this comparison, we report the average reprojection error with respect to one of the cameras, $C_0$. Equation~\ref{eq:reprerr} shows how this error is calculated with $^{\mathcal{W}}P$ as the generic joint expressed in the world reference system and $p^*$ as the corresponding ground truth :
\begin{equation}
\label{eq:reprerr}
e_\text{repr} = |p^* - K \cdot \mathcal{T}_{\mathcal{W}}^{C_0} \cdot ^\mathcal{W}P |
\end{equation}
Table~\ref{tab:results} shows the results we achieved. 
As depicted, the proposed method outperforms the baseline in all the cases: single-view, 2-camera network and 4-camera network. In the first two cases (single and 2-camera network) the improvement is from 50\% to 60\%, while, when multiple views are available, it is from 18\% to 32\%. It is also interesting to note that the most noisy joints are the ones relative to the legs as confirmed by other state-of-the-art works~\cite{cao2017realtime,insafutdinov2016deepercut,pishchulin2016deepcut}. 

\subsection{Implementation Details}
The system has been implemented and tested with Ubuntu 14.04 and Ubuntu 16.04 operating system using the Robot Operating System (ROS)~\cite{quigley2009ros} middleware. The code is entirely written in \texttt{C++} using the \texttt{Eigen}, \texttt{OpenCV} and \texttt{PCL} libraries.
\section{CONCLUSIONS AND FUTURE WORKS}
\label{sec:conclusions}

In this paper we presented a framework to compute the 3D body pose of each person in a RGB-D camera network using only its extrinsic calibration as a prior. The system does not make any assumption on the number of cameras, on the number of persons in the scene, on their initial poses or clothes and does not require the cameras to be synchronous. In our experimental setup we demonstrated the validity of our system over both single-view and multi-view approaches. In order to provide the best service to the Computer Vision community and to provide also a future baseline method to other researchers, we released the source code under the BSD license as part of the OpenPTrack library\footnote{\url{https://github.com/marketto89/open_ptrack}}. As future works, we plan to add a human dynamic model to guide the prediction of the Kalman Filters to further improve the performance achievable by our system (in particular for the lower joints) and to further validate the proposed system on a new RGB-Depth dataset annotated with the ground truth of the single links of the persons' body pose. The ground truth will be provided by a marker based commercial motion capture system.

\section*{ACKNOWLEDGEMENT}
This work was partially supported by U.S. National Science Foundation award IIS-1629302

%
%The authors would like to thank Jeff Burke and Randy Illum for the extensive test of our system performed at the UCLA Remap Group. 

\bibliography{mybib}{}
\bibliographystyle{ieeetr}

\end{document}